\documentclass[sigconf]{acmart}
\AtBeginDocument{%
  }

\setcopyright{acmlicensed}
\copyrightyear{2018}
\acmYear{2018}
\acmDOI{XXXXXXX.XXXXXXX}
\acmConference[Conference acronym 'XX]{Make sure to enter the correct
  conference title from your rights confirmation emai}{June 03--05,
  2018}{Woodstock, NY}
\acmISBN{978-1-4503-XXXX-X/18/06}

\usepackage{color}
\usepackage{graphicx}
\usepackage{textcomp}
\usepackage{xcolor}
\usepackage{url}
\usepackage{multirow}
\usepackage{times}
\usepackage{latexsym}
\usepackage{todonotes}
\usepackage{multirow}
\usepackage{graphicx}
\usepackage{amsmath}
\usepackage{amsfonts}
\usepackage{caption}
\usepackage[T1]{fontenc}
\usepackage{pifont}
\usepackage{subcaption}
\usepackage{booktabs}
\usepackage{colortbl}
\usepackage{tikz}
\usepackage{xcolor}

\definecolor{goldcolor}{RGB}{184,134,11}
\definecolor{lightbeige}{RGB}{245,245,220}
\definecolor{lightgray}{RGB}{211,211,211}
\definecolor{lightmint}{RGB}{176,224,230}
\definecolor{darkmint}{RGB}{96,168,154}

\usepackage{enumitem}
\usepackage{pifont}
\usepackage{graphicx}

\usepackage[utf8]{inputenc}

\definecolor{LightCyan}{rgb}{0.88,1,1}
\usepackage{graphicx}
\usepackage{subcaption}
\usepackage{enumitem}
\usepackage{xcolor}
\usepackage{color, colortbl}
\usepackage[utf8]{inputenc}
\usepackage{times}
\usepackage{latexsym}
\usepackage{comment}
\usepackage[utf8]{inputenc}
\usepackage{subcaption}
\usepackage[T1]{fontenc}
\usepackage{microtype}
\usepackage{algpseudocode}
\usepackage{times}
\usepackage{latexsym}
\usepackage{pgfplots}
\usepackage{svg}
\usepackage{makecell}
\usepackage{tikz}
\usetikzlibrary{positioning}
\usetikzlibrary{backgrounds}
\usepackage{tikzscale}
\usepackage{pgfplots}
\pgfplotsset{width=10cm,compat=1.9}
\usepgfplotslibrary{groupplots}
\usetikzlibrary{pgfplots.statistics}
\usetikzlibrary{patterns}
\usepackage{subcaption}
\usepackage{multirow}
\usepackage{adjustbox}
\usepackage{enumitem}
\usepackage{comment}
\usepgfplotslibrary{groupplots}
\usepackage[T1]{fontenc}
\definecolor{g1}{rgb}{0,0.8,0.4}
\definecolor{g4}{rgb}{0.88,1,0.88}
\definecolor{g2}{rgb}{0.66,1,0.66}
\definecolor{g3}{rgb}{0.8,1,0.8}

\definecolor{r1}{rgb}{1.0, 0.03, 0.0}
\colorlet{r2}{r1!50}
\colorlet{r3}{r1!30}
\colorlet{r4}{r1!15}

\begin{document}

\title{\texttt{PlotGen}: Multi-Agent LLM-based Scientific Data Visualization via Multimodal Feedback}







\author{Kanika Goswami}
\affiliation{%
  \institution{IGDTUW, Delhi}
  \country{India}}
\author{Puneet Mathur}
\affiliation{%
  \institution{Adobe Research}
  \country{USA}}
\author{Ryan Rossi}
\affiliation{%
  \institution{Adobe Research}
  \country{USA}}
\author{ Franck Dernoncourt}
\affiliation{%
  \institution{Adobe Research}
  \country{USA}}




\begin{abstract}

Scientific data visualization is pivotal for transforming raw data into comprehensible visual representations, enabling pattern recognition, forecasting, and the presentation of data-driven insights. However, novice users often face difficulties due to the complexity of selecting appropriate tools and mastering visualization techniques. Large Language Models (LLMs) have recently demonstrated potential in assisting code generation, though they struggle with accuracy and require iterative debugging. In this paper, we propose \texttt{PlotGen}, a novel multi-agent framework aimed at automating the creation of precise scientific visualizations. \texttt{PlotGen} orchestrates multiple LLM-based agents, including: (1) a Query Planning Agent that breaks down complex user requests into executable steps; (2) a Code Generation Agent that converts pseudocode into executable Python code; and three retrieval feedback agents—(3) a Numeric Feedback Agent, (4) a Lexical Feedback Agent, and (5) a Visual Feedback Agent—that leverage multimodal LLMs to iteratively refine the data accuracy, textual labels, and visual correctness of generated plots via self-reflection. Extensive experiments show that \texttt{PlotGen} outperforms strong baselines, achieving a 4-6\% improvement on the MatPlotBench dataset, leading to enhanced user trust in LLM-generated visualizations and improved novice productivity due to a reduction in debugging time needed for plot errors.
\end{abstract}



\maketitle

\section{Introduction}

\begin{figure*}
\centering
\scalebox{1}{
\includegraphics[width=1\textwidth]{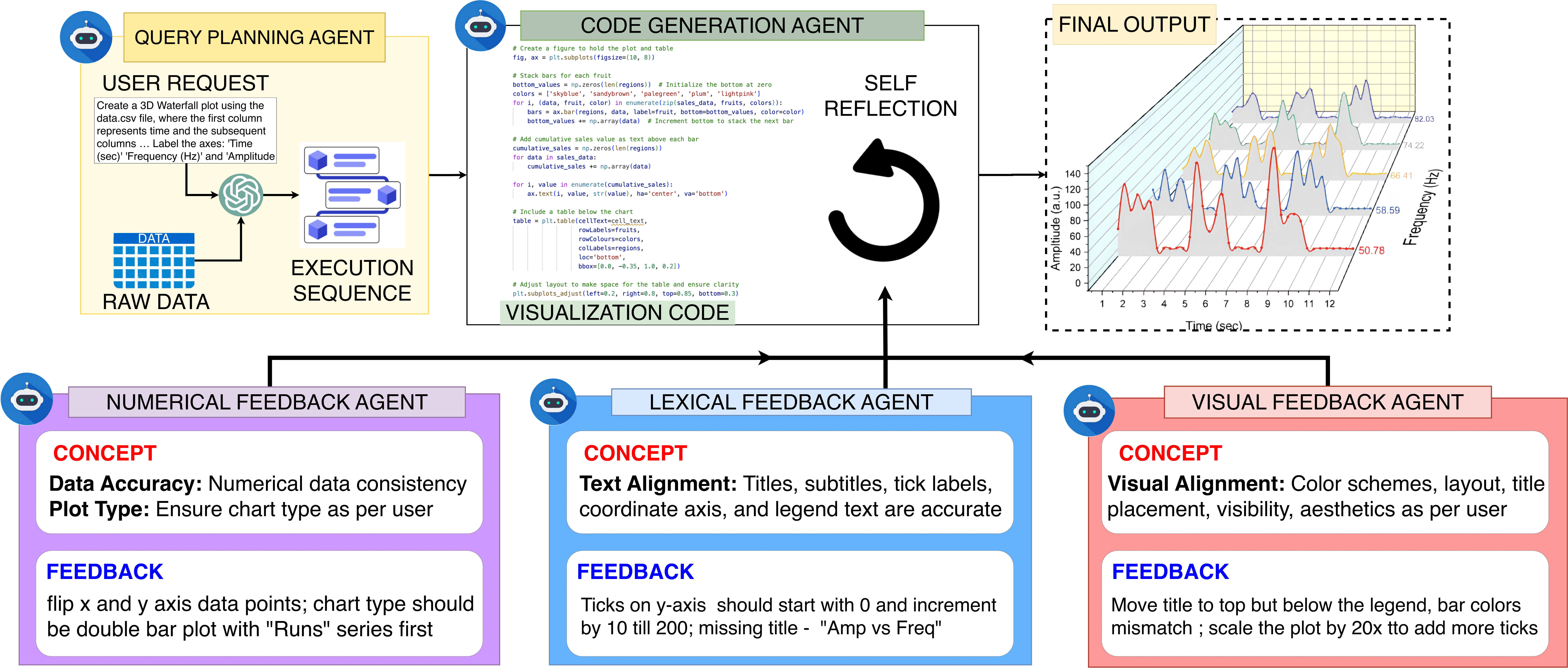}}
\caption{\small{\texttt{PlotGen} generates accurate scientific data visualizations based on user specifications by orchestrating multimodal LLMs: (1) Query Planning Agent that breaks down complex user requests into executable steps; (2) Code Generation Agent that converts pseudocode into executable Python code; and three code retrieval feedback agents—(3) Numeric Feedback Agent, (4) Lexical Feedback Agent, and (5) Visual Feedback Agent—that leverage multimodal LLMs to iteratively refine the data accuracy, textual labels, and visual aesthetics of generated plots via self-reflection.}}
\label{fig:main}
\end{figure*}

Scientific data visualization is essential for transforming raw data into visual representations, enabling the communication of complex information, the analysis of patterns in large datasets, and the formulation of evidence-based narratives. Professional data analysts and researchers have access to several sophisticated toolkits (e.g. Matplotlib, Seaborn and Plotly) to aid in the creation of diverse visualizations. However, novice users often face significant challenges when attempting to produce informative and accurate visualizations from raw data. These challenges stem from the difficulty of interpreting user requirements, choosing the appropriate visualization tool from the vast array of available options, and mastering the technical intricacies of graph plotting. Consequently, there is a growing need for more intuitive automated systems that support users with limited technical expertise to generate high-quality infographics.

Recently, Large Language Models (LLMs) have demonstrated remarkable capabilities in assisting a wide range of complex tasks, such as software development\cite{Qian2023ChatDevCA}, web navigation\cite{lai2024autowebglm}, and document editing\cite{mathur2024docpilot}, due to their proficiency in code generation\cite{Rozire2023CodeLO}, function-calling\cite{Qin2023ToolLLMFL}, tool utilization\cite{Patil2023GorillaLL, Schick2023ToolformerLM}, and self-reflective learning\cite{Shinn2023ReflexionAA,Ji2023TowardsML}. Recent advancements in LLM-based agents have opened new avenues for exploring their potential in automating the generation of scientific data visualizations based on user-defined queries, potentially lowering the barriers for novice users and enhancing accessibility to advanced data visualization techniques. Modern code-focused LLMs, such as CodeLlama\cite{Rozire2023CodeLO}, WizardCoder\cite{Luo2023WizardCoderEC}, and OpenAI GPT-4 have shown strong coding proficiency in Python to generate 2D and 3D chart figures via in-context learning. However, they struggle to faithfully adhere to user specifications due to hallucinations in code generation. AI-assisted data visualization often requires iterative refinement, where users critique intermediate plot outputs and adjust code generation to meet specific customization goals aligns with user expectations in terms of visual appearance, text label clarity, and precise data representation.

To address these challenges, we propose \texttt{PlotGen} (see Fig. \ref{fig:main}), a novel multi-agent framework designed to generate accurate scientific data visualizations based on user specifications by leveraging multimodal LLMs to iteratively debug the intermediate figures output by a code LLM via multimodal self-reflection. \texttt{PlotGen} consists of three unique multimodal feedback agents that provide sensory feedback—visual, lexical, and numerical—to the code generation agent, helping to rectify errors during the plotting process and better fulfill user queries through self-reflection. The system orchestrates multiple LLM agents, including: (1) a Query Planning Agent that decomposes complex user requests into a sequence of executable steps using chain-of-thought prompting; (2) a Code Generation Agent that transforms user data into a plot by converting pseudocode into executable Python code; (3) a Numeric Feedback Agent that ensures data rows and columns are accurately plotted and that the appropriate plot type is selected; (4) a Lexical Feedback Agent that verifies the accuracy of textual elements, such as titles, subtitles, axis labels, tick marks, and legend values, based on user requirements; and (5) a Visual Feedback Agent that checks that visual aspects of the chart, such as color schemes, layout, entity placement, and overall aesthetics, align with user specifications. Extensive experiments to benchmark the performance of our proposed \texttt{PlotGen} against strong baselines show that \texttt{PlotGen} improves the end-to-end task of scientific data visualization by 10-12\% across different LLM settings. Our \textbf{main contributions} are:

\begin{itemize}
    \item \textbf{\texttt{PlotGen}}, a \textbf{novel multi-agent LLM framework} that generates accurate scientific data visualizations based on user specifications by utilizing multimodal LLMs to iteratively refine intermediate plot outputs produced by a code LLM, via visual, lexical, and numerical self-reflection.

    \item Extensive experiments demonstrate that \textbf{\texttt{PlotGen} significantly outperforms strong baselines by 4-6\%} on the MatPlotBench dataset across various LLM configurations. Qualitative user studies further indicate that \texttt{PlotGen} enhances user trust in LLM-generated scientific visualizations and helps novice analysts increase productivity by reducing time spent debugging plotting errors.
\end{itemize}



\section{Related Work}


\noindent\textbf{Code LLMs}: With the advent of ChatGPT, several propriety LLMs like GPT-3.5, GPT-4, and Claude Sonnet-3.5 have emerged with increasingly strong code generation abilities. Moreover, numerous open-source LLMs such as CodeLlama \cite{roziere2023code}, DeepSeekCoder \cite{Guo2024DeepSeekCoderWT}, \cite{Luo2023WizardCoderEC}, \cite{Wei2023MagicoderSC} have also come out that are on par in producing executable code.

\noindent\textbf{LLM-based Data Visualization}: Several previous works have attempted to automate data visualization generation from natural language. \cite{Dibia2018Data2VisAG} was the first attempt to use LSTM to convert JSON data into Vega-Lite visualizations. \cite{dibia-2023-lida} explored use of LLMs to generate visualization code. Recent works studied the utility of LLMs like ChatGPT for generating charts from ambiguous natural language  \cite{Cheng2023IsGA,tian2024chartgpt}. \cite{berger2024visualization} expanded this line of work to include multimodal LLMs for chart plotting. \cite{xie2024haichart} tried to involve human feedback to refine the LLM generated plots via reinforcement learning. \cite{Yang2024MatPlotAgentMA} proposed a framework to provide visual feedback to LLMs for iterative refinement. Our work is different from existing works as it explores the use of multimodal feedback via LLM self reflection to resolve errors related to numeric values, lexical labeling and visual aesthetics.


\noindent\textbf{LLM Agents}: Recent years have seen a proliferation of frameworks that utilize Large Language Models (LLMs) to test their applications in practical scenarios \cite{Nakano2021WebGPTBQ, yao2022webshop, qin2023webcpm, zhou2023webarena}. The development of OpenAgents \cite{xie2023openagents} marked the introduction of an accessible platform that implements LLM-powered agents for daily use through three specialized components: Data Agent, Plugins Agent, and Web Agent. A groundbreaking simulation system that replicates human behavioral patterns was developed by \cite{park2023generative}, enabling software agents to computationally reproduce authentic human actions and interactions. In the gaming realm, Voyager \cite{Wang2023VoyagerAO} emerged as the pioneer LLM-controlled autonomous agent within Minecraft, engineered to continuously discover its surroundings, develop diverse abilities, and generate novel discoveries without human intervention. The software development sphere saw innovation through ChatDev \cite{Qian2023ChatDevCA}, which established a virtual software company operated through chat interfaces and adhering to waterfall development principles. Building upon these advances, our research investigates how LLM-based agents can contribute to scientific data visualization, an essential domain for modern researchers.

\section{Methodology}


\noindent\textbf{Task Description}: Given a user request $r$, specifying the visualization requirements including chart type, spatial arrangement of chart entities, and aesthetic
preferences, along with a data table $D = \{d_{1,1}, d_{1,2}...d_{nxm}\}$ as inputs, the scientific data visualization task should output a figure $v$. We propose \texttt{PlotGen} to automate this challenging task that orchestrates a multi-agent LLM framework consisting of five agents as follows:

\noindent\textbf{(1) Query Planning Agent}: Query Planning Agent uses an LLM to breakdown a complex user request into a sequence of executable steps via chain-of-thought prompting. Each step in the thought chain corresponds to an explicit instruction to the code LLM responsible for generating the visualization. It highlights the programming language required and intermediate programming steps that specify the function calls, parameters, and return types. The output also instructs the code LLM on the data file and its format provided for visualization, along with the visual characteristics desired by the user.

\noindent\textbf{(2) Code Generation Agent}: The Code Generation Agent is the most crucial component of the framework as it is responsible for generating an error-free executable code that can transform the user provided data into a scientific visualization as per user request. We use specific code LLMs such as GPT-3.5, GPT-4o for this purpose. However, code LLMs are prone to programming errors that may hinder the final plot generation. Hence, we include a self-reflection step in the code generation agent by iterating on the debugger error response in case of failed code execution. The debugger error message is passed as feedback to the code LLM to fix issues related to syntax, library imports, function arguments, and data formatting. To prevent an infinite loop of iteration over coding errors, we restrict the number of iterations of self-debugging.

\noindent\textbf{Multimodal Feedback}: Humans tend to repeatedly refine their drafts to reach satisfactory outputs based on sensory feedback. We hypothesize that the code generation agent should also have the ability to receive external feedback to rectify mistakes during the plotting process to better fulfill the user’s queries. Some of the errors may be difficult to diagnose in code but become apparent when observed visually through "eyes", reading out the textual phrases, or asking probing questions about the data points. To this end, we introduce \textit{Multimodal Feedback Agents} that provide visual, lexical, and numerical feedback to the code agent for self-reflection and improving the final figure output.

\noindent\textbf{(3) Numeric Feedback Agent}: 
The numeric agent is responsible for ensuring the accuracy of underlying data in the figure by ensuring all data rows and columns are appropriately plotted and the right kind of plot has been selected for visualization. Numeric feedback agent uses GPT-4V to de-render the draft figure to get back the underlying data from the plot. It then compares the de-rendered data with the original data set to check for any discrepancies between the two. If the figure has the correct plot, the two data collections should show similar numeric trends across rows and columns even if the corresponding numerical values are not exactly same. Further, the numeric feedback agent predicts if the draft figure type matches the user's expectation. In case the data trends don't match or the plot type is inconsistent with the user request, the agent provides text feedback to the code agent to fix the error by refining the code used to generate the plot. 

\noindent\textbf{(4) Lexical Feedback Agent}: Chart figures contain labels that help comprehend the plotted data. The textual markers corresponding to these labels are passed as arguments during the code generation step. The lexical feedback agent is responsible for ensuring the textual markers such as titles, subtitles, tick labels, axis descriptions, and legend values are accurate as per the raw data and user requirements. Lexical feedback agent uses GPT-4V to "read" and verify the textual values of chart labels with the corresponding ground truth values present in the data sheet. In case the chart labels don't match, it provides feedback to the code generation agent to fix them in the next iteration.

\noindent\textbf{(5) Visual Feedback Agent}: Visual feedback agent is responsible to ensure that visual aspects in chart figures such as color schemes, layout, placement of chart entities, and aesthetics are aligned as per user requirements. To this end, visual feedback agent uses GPT-4V to observe the draft figure similar to how a human "sees" the figure and provides textual feedback to the code generation agent for draft refinement. 

\noindent All three of the feedback agents are allowed to provide their respective feedback sequentially. We iterate each feedback agent repeatedly until satisfactory results are achieved in that domain or the maximum trial count runs out.

\section{Experiments}


\noindent\textbf{Multimodal LLMs}: We experiment with GPT-4V \cite{Achiam2023GPT4TR} for multimodal feedback agents.

\noindent\textbf{Code LLMs}: For code generation agent in \texttt{PlotGen}, we experiment with both closed source (GPT-3.5, GPT-4) as well as open source LLMs (Magicoder-SDS-6.7B \cite{Wei2023MagicoderSC} and WizardCoder-Python33B-V1.1 \cite{Luo2023WizardCoderEC}) . We set the decoding temperature to 0 for all code LLMs and use their respective APIs for closed source LLMs.

\noindent \textbf{Dataset}: We use MatPlotBench \cite{Yang2024MatPlotAgentMA} as benchmark dataset to assess the effectiveness of our proposed framework. This corpus contains 100 high-quality user queries and raw data mapped to ground-truth visualizations figures close to real-world scenarios.

\begin{table*}[ht]
\centering
\scriptsize
\begin{tabular}{l|c|c|c|c}
\toprule
\bf Method & \bf WizardCoder-33B & \bf Magicoder-6.7B & \bf GPT-3.5 & \bf GPT-4 \\
\hline
Dir. Decoding & 36.94 & 38.49 & 38.03 &  48.86 \\
0-shot CoT & 35.81 & 37.95 & 37.14  & 45.42 \\
MatplotAgent & 45.96 & 51.70 & 47.51 & 61.16 \\ \hline
\rowcolor{g3} \texttt{PlotGen} & \bf 48.82 & \bf 57.13 & \bf 53.25 & \bf 65.67 \\
\bottomrule
\end{tabular}
\caption{\label{tab:main_results} Performance of \texttt{PlotGen} compared with baselines on MatPlotBench.}
\end{table*}

\begin{figure*}[ht]
    \centering
    
    \includegraphics[width=0.5\textwidth]{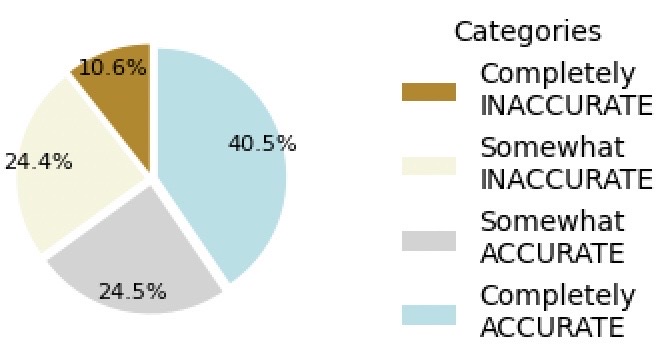}
        \label{fig:plot2}
\caption{\label{tab:ablation_results}User Evaluation}
\end{figure*}

\noindent \textbf{Baselines}: We benchmark \texttt{PlotGen} against several strong baselines: \textbf{(1) Direct Decoding}: The LLM directly generates the code for rendering the visualization. \textbf{(2) Zero-Shot Chain-of-thought prompting}: The LLM is prompted to inference with the zero-shot chain of thought mechanism based on the user description. \textbf{(3) MatPlotAgent}: We benchmark against MatPlotAgent \cite{Yang2024MatPlotAgentMA} which utilizes GPT-4 as code LLM and GPT-4V for visual feedback. We experiment all baselines with various code LLM backbones - GPT-3.5, GPT-4, Magicoder-SDS-6.7B \cite{Wei2023MagicoderSC}, and WizardCoder-Python33B-V1.1 \cite{Luo2023WizardCoderEC}.

\noindent\textbf{Evaluation}: Following \cite{Yang2024MatPlotAgentMA}, we use LLM-as-a-judge automatic scoring metric between 0 to 100 to evaluate model-generated visualizations with corresponding ground-truth as a reference.  The authors show that automatic evaluation scores provided by GPT-4V are sufficiently reliable due to their strong correlated with human evaluation results.

\noindent\textbf{Libraries}: We use MatPlotlib to generate final output from LLM code output. We use Huggingface library to run open-source code LLMs.


\section{Results and Discussion}

\noindent\textbf{Performance Evaluation}: Table~\ref{tab:main_results} compares the performance of \texttt{PlotGen} with baseline methods on the MatPlotBench dataset. We observe that \texttt{PlotGen} significantly outperforms strong baselines across both open-source and closed source code LLMs. The direct decoding methods show degraded performance due to their inability to correct common plotting mistakes in text labels and visual appearance. We observe that naive zero-shot CoT mechanism does not significantly enhance the performance of code LLMs. While MatPlotAgent shows better performance, it struggles to handle complex charts that require fine-grained numerical accuracy and precise visual-textual alignment of chart labels with plot components. \texttt{PlotGen} effectively captures user requirements, translates them into error-free code, and reduces errors caused by the lack of multimodal perception in baseline methods. \texttt{PlotGen} shows best performance with GPT-4 backbone, where inclusion of lexical, numerical, and visual feedback helps it to recover from various concurrent errors during the plotting process, demonstrating the effectiveness of our multi-agent approach. 

\noindent\textbf{Ablation Study}: We conduct an ablation study to verify the effectiveness of each feedback agent in \texttt{PlotGen}. Ablation of the Lexical Feedback agent causes 5-7\% performance deterioration as it is unable to iteratively refine textual details such as titles, subtitles, tick labels, and legend text, which is crucial for clarity in data interpretation. The Numerical Feedback agent helps reduce ambiguity in data  plotting by verifying the underlying data trends and corresponding chart types. Its usefulness is particularly pronounced in co-located charts with complex legend hierarchies that are prone to mispredictions due to LLM hallucinations.  We observed a severe performance drop (10-15\%) across all code LLMs in the absence of the Visual Feedback agent, which plays a critical role in maintaining visual aesthetic quality as per user requirements. Without the visual feedback, \texttt{PlotGen} runs into common plotting mistakes related to color schemes, scaling, layout, tick labels and legend mapping. 

\noindent\textbf{User Evaluation}: Five participants evaluated 200 randomly sampled user requests from the MatPlotBench dataset to study the usefulness and accuracy of the plots generated by \texttt{PlotGen}. The evaluation results demonstrated strong positive reception, with participants rating the attributions as \textit{Completely Accurate} ($40.5\%$) or \textit{Somewhat Accurate} ($24.5\%$) for natural language based plot generation in PlotGen. Plots were found to be more "Completely Inaccurate" in $10.6\%$ of the cases. Participants described the outputs useful in reducing time spent debugging plotting errors.

\section{Conclusion}

In this paper, we introduced \texttt{PlotGen}, a novel multi-agent framework designed to automate the process of scientific data visualization through a multimodal self-reflection. By coordinating multiple LLM-based agents responsible for query planning, code generation, and multimodal feedback, \texttt{PlotGen} addresses the challenges faced by novice users in generating accurate visualizations according to user preferences. \texttt{PlotGen} significantly improves the quality of LLM-generated plots by 10-12\% compared to strong baselines, enhancing user trust, reducing debugging time, and improving accessibility. Future work will extend \texttt{PlotGen} beyond traditional table data visualization to explore its application in real-time environments such as interactive dashboards, Virtual Reality simulations and visual arts.

\bibliographystyle{ACM-Reference-Format}
\bibliography{sample-base}


\end{document}